%% file: main.tex
\pdfoutput=1
\documentclass[10pt,twocolumn,letterpaper]{article}

\usepackage{multirow}
\usepackage{graphicx}
\usepackage{adjustbox}
\usepackage{caption}
\usepackage{blindtext}
\usepackage{booktabs}
\usepackage{authblk}

\usepackage[pagenumbers]{cvpr} 

\input{preamble}

\definecolor{cvprblue}{rgb}{0.21,0.49,0.74}
\usepackage[pagebackref,breaklinks,colorlinks,citecolor=cvprblue]{hyperref}

\def\paperID{9172} 
\def\confName{CVPR}
\def\confYear{2024}

\title{SPDiffusion: Semantic Protection Diffusion Models for Multi-concept Text-to-image Generation}

\author {
    Yang Zhang\textsuperscript{\rm 1,\rm 4}, 
    Rui Zhang\textsuperscript{\rm 1,\rm 4}\thanks{Corresponding author.},
    Xuecheng Nie\textsuperscript{\rm 2}, 
    Haochen Li\textsuperscript{\rm 3,\rm 4},
    Jikun Chen\textsuperscript{\rm 1,\rm 4},
    Yifan Hao\textsuperscript{\rm 1,\rm 4},
    Xin Zhang\textsuperscript{\rm 1,\rm 4},
    Luoqi Liu\textsuperscript{\rm 2}, 
    Ling Li\textsuperscript{\rm 3,\rm 4} \\
    \textsuperscript{\rm 1}State Key Lab of Processors, Institute of Computing Technology, CAS\\
    \textsuperscript{\rm 2}MT Lab,Meitu Inc. \\
    \textsuperscript{\rm 3}Intelligent Software Research Center, Institute of Software, CAS\\
    \textsuperscript{\rm 4}University of Chinese Academy of Sciences\\
}

\begin{document}
\maketitle
\input{sec/0_abstract}
\input{sec/1_intro}

\input{sec/2_related_work}
\input{sec/3_method}
\input{sec/4_expriment}
\input{sec/5_conclusion}
{
    \small
    \bibliographystyle{ieeenat_fullname}
    \bibliography{main}
}

\input{sec/X_suppl}

\end{document}

%% file: preamble.tex
\usepackage[dvipsnames]{xcolor}


%% file: sec/0_abstract.tex
\begin{abstract} Recent text-to-image models have achieved impressive results in generating high-quality images. However, when tasked with multi-concept generation creating images that contain multiple characters or objects, existing methods often suffer from semantic entanglement, including concept entanglement and improper attribute binding, leading to significant text-image inconsistency. We identify that semantic entanglement arises when certain regions of the latent features attend to incorrect concept and attribute tokens. In this work, we propose the Semantic Protection Diffusion Model (SPDiffusion) to address both concept entanglement and improper attribute binding using only a text prompt as input.
The SPDiffusion framework introduces a novel concept region extraction method SP-Extraction to resolve region entanglement in cross-attention, along with SP-Attn, which protects concept regions from the influence of irrelevant attributes and concepts.
To evaluate our method, we test it on existing benchmarks, where SPDiffusion achieves state-of-the-art results, demonstrating its effectiveness. \end{abstract}

%% file: sec/1_intro.tex
\begin{figure}[t]
    \centering
    \includegraphics[width=0.47\textwidth]{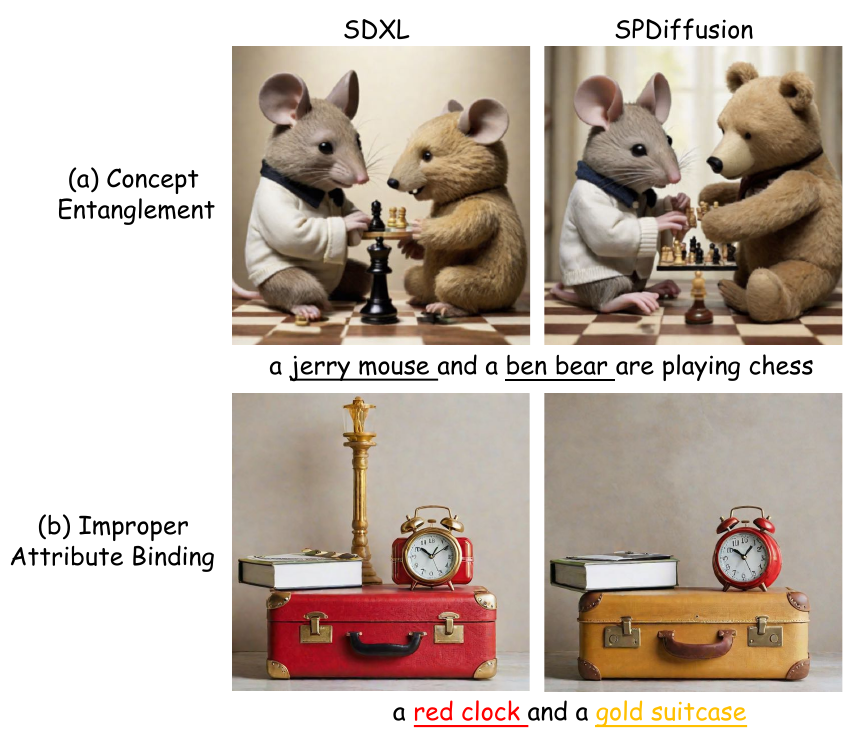} 
    \caption{
    \textbf{Semantic Entanglement}. Existing diffusion models usually suffer from semantic entanglement problem in multi-concept text-to-image generation, which contains following sub-problems:
    (a). Concept Entanglement. One concept feature transfers to another concept. (e.g., bear exhibit mouse like ear and mouth.)
    (b). Improper Attribute Binding. attribute of one concept binds to another concept. (e.g., red color binds to suitcase and gold color binds to clock. )
    }
    \label{fig:problem_def}
\end{figure}

\section{Introduction}
\label{sec:intro}
Recent text-to-image diffusion models, such as DALLE \cite{dalle}, Stable Diffusion \cite{sd}, and PixArt-alpha \cite{chen2024pixart}, have demonstrated impressive capabilities in generating realistic images from text prompts, facilitating applications such as story illustration \cite{zhou2024storydiffusion} and portrait creation \cite{li2024photomaker}. 
However, these models are primarily adept at producing single-concept images, those with a single character or object. 
When tasked with generating multi-concept images, they frequently encounter semantic entanglement issues including concept entanglement and improper attribute binding. 
As shown in Fig.\ref{fig:problem_def} (a). concept entanglement: One concept transfers to another concept. (b). improper attribute binding: attribute of one concept binds to another concept. 

Most existing methods aim to address the issue of improper attribute binding. 
Several methods \cite{chefer2023attend, li2023divide, meral2024conform, rassin2024linguistic} enhance text-image alignment by optimizing latent representations via repeated backpropagation during inference. However, this can shift the latent space away from the real image distribution, thereby reducing image quality. Furthermore, repeated backpropagation increases inference time. 
Other approaches \cite{zhu2024isolated, feng2022training, liu2022compositional} split the prompt and process each part separately with diffusion model, yet this makes it challenging to generate coherent and natural synthesis results. 
\cite{zhuang2024magnet} addresses attribute binding by reinforcing the association between attributes and concepts within the text encoding space but still faces concept entanglement issues.
\cite{dahary2024yourself} mitigates concept entanglement by restricting subjects within bounded boxes, though this approach requires additional layout inputs.

Cross-attention map and self-attention map are two of most important components in diffusion models, since cross-attention map \cite{hertz2022prompt} describes the feature merging relation between image feature and the text feature and self-attention map \cite{tumanyan2023plug} describes how image feature produces. Analyzing the cross-attention map, we find that semantic entanglement occurs when one concept token attends to multiple concept region or attribute attends to incorrect concept region. This incorrect feature merging relation leads to incorrect image feature producing. As shown in Fig.\ref{fig:SPA_motivation}(a), mouse token attends in two regions in cross-attention map with SDXL \cite{podell2023sdxl}, thus in self-attention map, bear region queries mouse ear in red box producing a mouse like ear. Therefore, in order to generate correct concept and its attribute, we need to obtain regions of concept from a incorrect image generation process and eliminate the attention of region of concept to irrelevant tokens by elaborately constraining the cross-attention map. 
Although cross-attention map contains the region of concepts\cite{hertz2022prompt}, extracting region of concept from incorrect image generation process is not easy, as the region is scattered and overlaps irrelevant concept. By analyzing different threshold, we find that high threshold value can filter out the irrelevant region in cross attention map. This motivates us to extract concept regions from both the cross-attention and self-attention maps.

In this work, we propose SPDiffusion, a novel training-free multi-concept text-to-image generation method that uses only text prompts as input to address semantic entanglement. 
The key ideas of SPDiffusion are extracting the regions of concepts in semantic entanglement generation process and protecting the semantic of the region from being confused with other non-corresponding attributes or concepts. 
In the SPDiffusion framework, we propose a novel SP-Extraction method to extract concept region from incorrect generation process, which extract anchor point of a concept from cross-attention map and extract real region of the concept from self-attention map by filtering high attention regions to the anchor point. 
With the regions of concepts, we propose the SP-Mask to indicate which irrelevant concept and attribute tokens should be masked for a concept region. Furthermore, we propose the SP-Attn to protect the concept region from merging irrelevant attribute and concept features with SP-Mask. 
SPDiffusion is capable of significantly mitigating the semantic entanglement problem without extra layout input. 

We evaluate our method on CC-500 \cite{feng2022training} dataset and other two datasets Wearing-100 and Animals-100 designed for better semantic entanglement evaluation.
Our method outperforms other baselines \cite{feng2022training,liu2022compositional,zhuang2024magnet} in BLIP-VQA \cite{huang2023t2i} on all three datasets, which achieves state-of-art results. We also use InternVL \cite{chen2024internvl}, a large visual language model, for more accurate scoring, which confirms our method’s efficiency in semantic entanglement problems.

Our contributions are summarized as follows:

1. We propose a new framework that addresses both concept entanglement and improper attribute binding issues without the need for additional layout input.

2. We introduce a novel region extraction technique for handling semantic entanglement in diffusion model process. 

3. Experimental results show that our method outperforms baseline methods in addressing semantic entanglement problems.

%% file: sec/2_related_work.tex
\begin{figure}[t]
    \centering
    \includegraphics[width=0.47\textwidth]{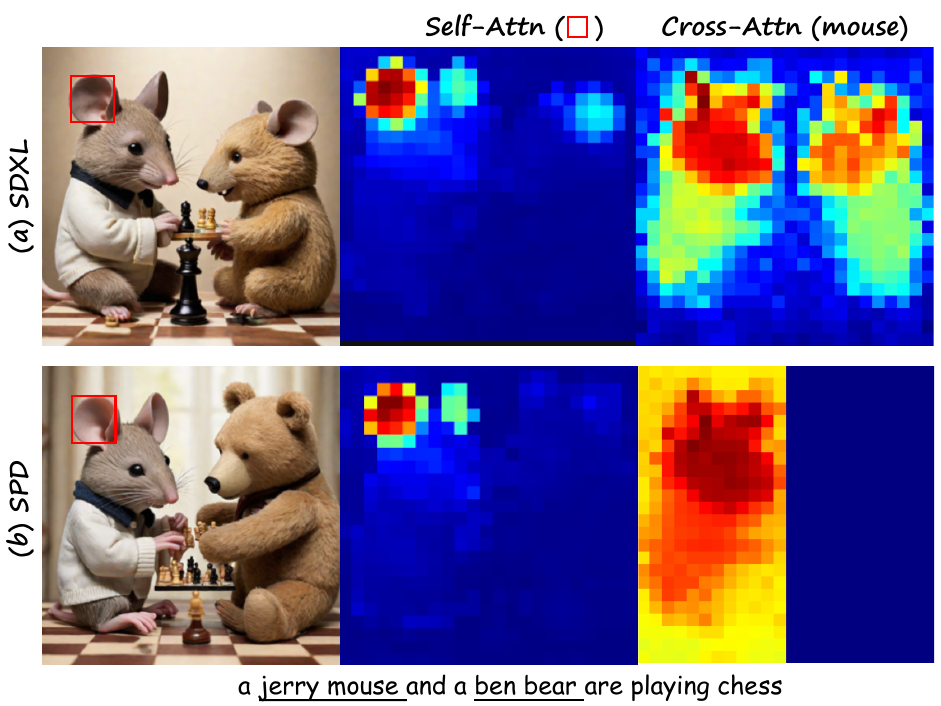} 
    \caption{
    \textbf{Semantic Entanglement Visualization}. 
    (a) Cross-attention map visualization shows both the mouse and bear regions merging mouse features, causing the bear's ear region to query image features (highlighted in the red box) associated with the mouse, resulting in a mouse-like ear.
    (b) When the bear region does not merge mouse features, it does not query the mouse ear feature in the red box, maintaining distinct bear features.
    }
    \label{fig:SPA_motivation}
\end{figure}
\section{Related work}
\label{sec:Related work}
\subsection{Text-to-image diffusion models}
Text-to-image diffusion models \cite{sd,dalle3,chen2024pixart,podell2023sdxl} have become the most popular image generative models. They are trained in large image-text pair datasets \cite{schuhmann2022laion} and can generate high quality and diverse images with only text as input. Since the text prompt in the training datasets are mostly describing only one concept and its attribute, the text-to-image diffusion models often suffer semantic entanglement problems, which one concept appearance entangles with another or attribute of one concept binds to another concept.

\subsection{Semantic Entanglement}
The semantic entanglement usually contains two sub-problems, concept entanglement and improper attribute binding. Concept entanglement refers to one concept appearance entangled with another concept and improper attribute binding means one concept's attribute binds to another concept.

\subsubsection{Concept Entanglement}
Several methods \cite{chen2024training, battash2024obtaining, endo2024masked, xiao2024fastcomposer} address concept entanglement by supervising the cross-attention map to align with a given input layout box, while \cite{kim2023dense} uses attention modulation to achieve this alignment. \cite{dahary2024yourself} supervises both cross-attention and self-attention maps to align with the input layout box, guiding it to focus on specific concepts and attributes. However, all of these approaches rely on additional layout box input, which can be inconvenient. Approaches such as \cite{kong2025omg, kwon2024concept} generate a layout image first, then separately generate concepts and weight the predicted noise with detected masks using SAM\cite{kirillov2023segment}. As this involves two complete denoising processes, it significantly increases inference time and depends on an extra segmentation model.

\subsection{Improper Attribute Binding}
To address improper attribute binding, various methods have been introduced. 
Various method \cite{chefer2023attend,li2023divide, meral2024conform, rassin2024linguistic, wang2024tokencompose, battash2024obtaining,zhang2025object} supervise attention maps during inference by using backpropagation to identify optimal latent representations. However, directly modifying latent representations can push the latent space out of distribution, resulting in quality degradation, while multiple backpropagation iterations significantly increase inference time. 
As diffusion models generate relatively accurate semantic alignment for single-concept images, many approaches \cite{zhu2024isolated, feng2022training, liu2022compositional} have attempted to handle multi-concept generation separately and combine the separate generation results.
However, prompt splitting complicates the ability of partial regions to capture the full semantic context of the input prompt, leading to potential semantic loss. Additionally, generating concepts separately significantly increases inference time, scaling linearly with the number of concepts.
Magnet \cite{zhuang2024magnet} strengthens the connection between concepts and their attributes within the text encoder space; however, it struggles to associate attributes with concepts that have strong attribute biases. OABinding \cite{trusca2024object} identifies concept regions from cross-attention maps and restricts the focus of these regions to their specific attributes. Our method differs in two key ways: 

(1). We mask only irrelevant concepts and attributes, preserving shared and global descriptions, while OABinding has difficulty managing shared attributes. 

(2). OABinding focuses solely on attribute binding, which may fail to isolate concept regions effectively when concept entanglement occurs in the cross-attention map.

%% file: sec/3_method.tex
\section{Motivation}
\begin{figure}[t]
    \centering
    \includegraphics[width=0.47\textwidth]{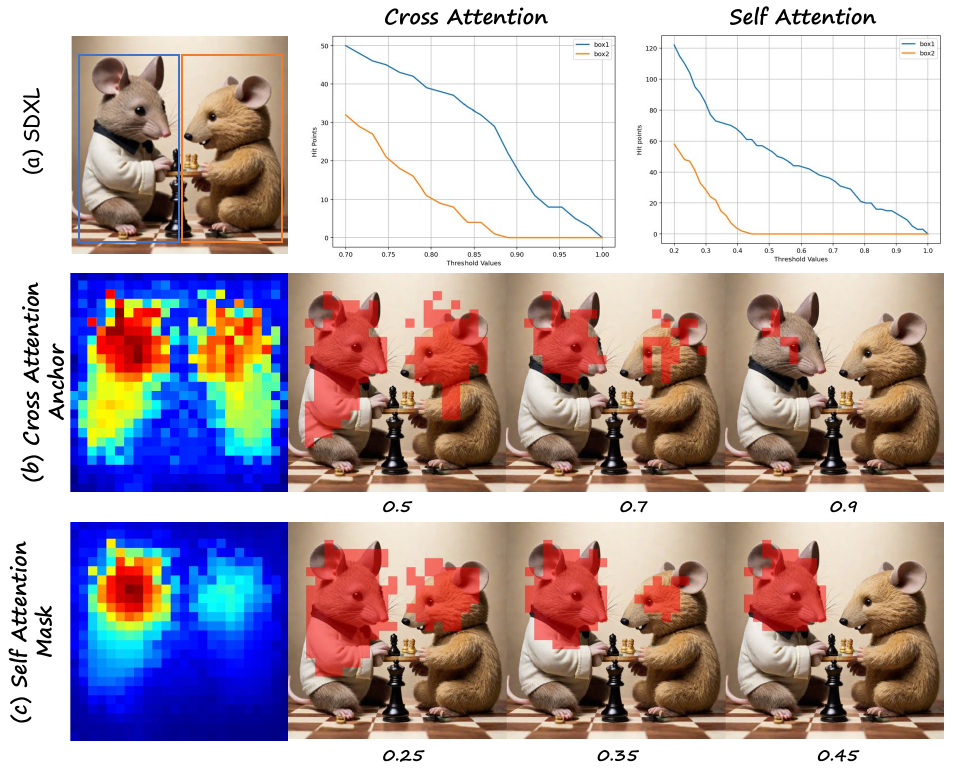} 
    \caption{
    \textbf{Concept Region Extraction.} 
        We visualize the normalized heat maps of both the cross-attention of mouse token and self-attention maps of anchor points. Additionally, we display masks and points within blue and orange boxes under varying thresholds.
    }
    \label{fig:SPM_motivation}
\end{figure}

\begin{figure*}[t]
    \centering
    \includegraphics[width=1\textwidth]{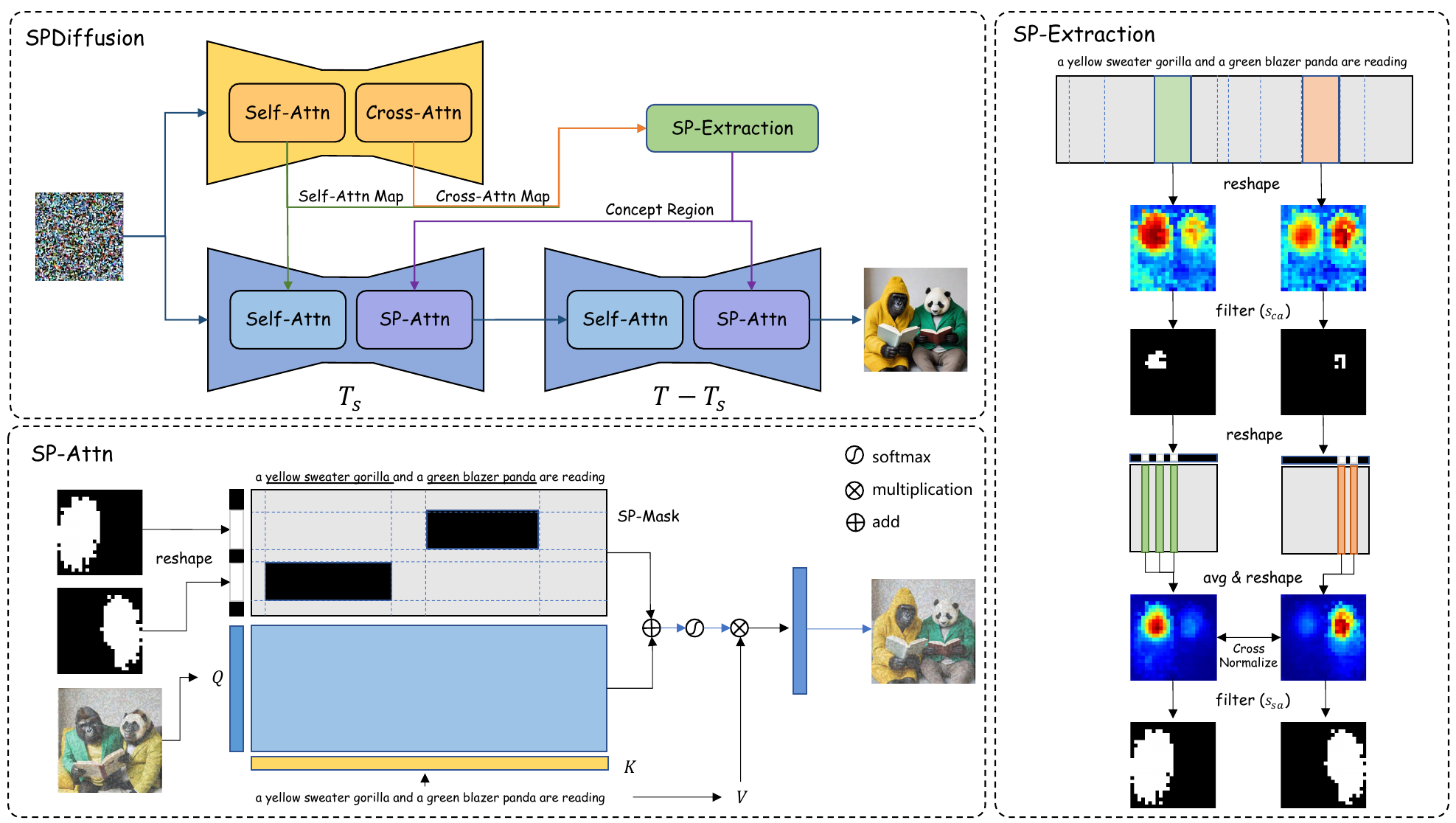} 
    \caption{
    \textbf{Overview of SPDiffusion} 
    }
    \label{fig:method}
\end{figure*}

\subsection{Semantic Entanglement}
The cross-attention map and self-attention map are two of the most important component in diffusion models. Previous work \cite{hertz2022prompt,tumanyan2023plug} shows cross-attention map controls how image feature merges text embeddings and self-attention contains information how image feature query other image feature to produce new features. To address the semantic entanglement problem, we analyzed the self-attention and cross-attention in SDXL 
 \cite{podell2023sdxl}. We observe that concept region attends to incorrect attribute or concept in cross-attention map in SDXL \cite{podell2023sdxl}. Since image feature queries other image features containing similar semantics in self-attention, one concept region may query features of another concept, leading to an entangled appearance.
As illustrated in Fig.\ref{fig:SPA_motivation}, the third column visualizes cross-attention map of mouse token and the second column visualizes self-attention map of red box region. 
In Fig.\ref{fig:SPA_motivation}(a), the bear region incorporates mouse embedding in cross-attention in SDXL\cite{podell2023sdxl}. Therefore, the bear region queries mouse ear features in red box, resulting in a bear with mouse-like ears. 
In Fig.\ref{fig:SPA_motivation}(b), when the bear region excludes mouse embedding in cross-attention, it does not query the mouse ear region in self-attention, maintaining distinct bear features. 
Thus, we aim to eliminate the phenomenon where concept regions merge irrelevant attribute and concept features by masking the irrelevant tokens in the cross-attention map.

\subsection{Concept Region Extraction}
\label{sec:concept area}
Extensive prior works \cite{hertz2022prompt, chefer2023attend, li2023divide} have shown that cross-attention maps contain concept region information. However, these regions in the cross-attention map are often fragmented and inaccurate, especially in cases of semantic entanglement. In Fig.\ref{fig:SPM_motivation}(b), we visualize the cross-attention map of mouse token and use different threshold to filter the points. we find that normal threshold filters two concept region, and although high threshold filters one concept region, the area is not large enough to be a mask. In Fig.\ref{fig:SPM_motivation}(a), We also count the number of points that fall within the blue and orange box. We can observe that the points in blue box are always more than orange box, which means the the mouse region pay more attention to mouse token.
In Fig.\ref{fig:SPM_motivation}(c), we visualize the self-attention map of the points in second row and forth column and find that the area is clear and distinct. 
Based on the analyze above, we can use high threshold to get anchor points of concept in cross-attention and get actual mask of concept in self-attention to handle the entangled concept region in cross-attention map. 

\section{Method}
\subsection{Preliminaries}
Diffusion models \cite{ho2020denoising,song2020denoising} have recently become the most popular generative models. Given a noisy image $ x_t $, a diffusion model $ \epsilon_\theta $ (usually a neural network) will predict the noise $ \epsilon_{t} $ present in the image and subtracts it from $ x_t $ to obtain $ x_{t-1} $.
This process will be repeated $ T $ times, starting from a random noise $ x_T $ and ultimately producing a completely denoised image $ x_0 $.
Stable Diffusion \cite{sd} utilizes an autoencoder to represent an image $ x_t $ as a latent $ z_t $ with significantly smaller width and height.
Noise prediction during both training and inference occurs in the latent space with diffusion model $ \epsilon_\theta $, which significantly reduces computational resources and inference time.

For the diffusion model structure $ \epsilon_\theta $, we use Stable Diffusion XL \cite{podell2023sdxl} as an example. The model employs a U-Net \cite{ronneberger2015u} as its backbone, which includes multiple layers of transformer blocks \cite{transformer}, typically consisting of Self-Attention (Self-Attn), Cross-Attention (Cross-Attn), and Feed-Forward Networks (FFN).

Before feeding into the $l$-th Transformer block, the intermediate features $\phi^{(l)}(z_t)$ are produced by the previous layers from $z_t$. This $\phi^{(l)}(z_t)$ is then projected to $Q_t^{(l)}$, $K_t^{(l)}$, and $V_t^{(l)}$ through linear projections $f_Q^{(l)}$, $f_K^{(l)}$, and $f_V^{(l)}$, respectively. The text embedding $C(p)$, where $C(\cdot)$ is the text encoder and $p$ is the text prompt, is similarly projected to $K_t^{(l)}$ and $V_t^{(l)}$ via $f_K^{(l)}$ and $f_V^{(l)}$. The attention map is then obtained by the following formulation:

\begin{equation}
  A_t^{(l)} = \text{Softmax}\left(\frac{Q_t^{(l)} K_t^{(l)^{T}}}{\sqrt{d}}\right),
  \label{eq:attention map}
\end{equation}

where $d$ represents the dimension of $Q$. By multiplying the attention map $A_t^{(l)}$ by $V_t^{(l)}$ and projecting back through the linear projection $f_{out}^{(l)}$, we obtain the updated intermediate features $\phi^{(l)\prime}(z_t)$:

\begin{equation}
  \phi^{(l)\prime}(z_t) = f_{out}^{(l)} \left(A_t^{(l)} V_t^{(l)}\right).
  \label{eq:hiddenstates}
\end{equation}

\subsection{SPDiffusion} 
In this work, we propose a novel method, SPDiffusion, to address semantic entanglement problems using only text prompts as input. As shown in Fig.\ref{fig:method}, SPDiffusion contains two main components, SP-Extraction and SP-Attn. SP-Extraction extract concept region from cross-attention and self-attention in a normal denoising process. SP-Attn use the concept region to protect the concept region from the influence of irrelevant attributes and concepts. We first introduce SP-Extraction and SP-Attn, followed by an overview of the entire framework.

\subsubsection{SP-Extraction}
\label{sec:Concept Region Extraction}
The core idea of SPDiffusion is to protect concept regions from the influence of irrelevant attributes and concepts. While previous work \cite{dahary2024yourself} relies on additional layout inputs to define concept region, our approach enables the diffusion model to determine concept positions autonomously, avoiding the need for external object detectors. Prior work \cite{trusca2024object} uses cross-attention maps to identify concept regions in cases of attribute binding errors; however, as analyzed in Sec.\ref{sec:concept area}, these regions are often imprecise due to incorrect attention in cross-attention. Additionally, concept region derived from cross-attention are often dispersed across a broad scope, leading to overlap between concept regions. In our approach, we use cross-attention to obtain anchor points for concepts and self-attention to create more accurate concept masks. Furthermore, we apply cross-normalization to reduce the impact of incorrect attention, allowing for more robust thresholding.

Formally, given a prompt $p$, we use an NLP library (e.g., spaCy\cite{spacy2}) to extract concepts $\mathbb{E}=\{e_{1}, e_{2}, \dots, e_{n}\}$ and their attributes $\mathbb{A}=\{a_{1}, a_{2}, \dots, a_{n}\}$. Whin a denoising process, we aim to obtain the regions of these concepts, $\mathbb{D}=\{d_{1}, d_{2}, \dots, d_{n}\}$, decided by the diffusion model itself. We aggregate the cross-attention maps across selected steps and layers to produce averaged results, using min-max normalization to rescale values to the range $[0,1]$:

\begin{equation}
\label{eq:CMc1}
    \bar{A}_{ca} = \text{MinMaxNorm}\left(\frac{1}{T} \cdot \frac{1}{L} \sum_{t} \sum_{l} A^{l}_{ca(t)}\right),
\end{equation}

where $T$ and $L$ denote the numbers of selected steps and layers, respectively. We then determine anchor points for each concept by applying a relatively high threshold value $s_{ca}$:

\begin{equation}
\label{eq:CMc2}
    m_k[i] = 
    \begin{cases}
    1, & \bar{A}_{ca}^{e_k}[i] \geq s_{ca} \\
    0, & \text{otherwise} 
    \end{cases}, 1\leq i \leq w \times h  ,
\end{equation}

where 

\begin{equation}
\label{eq:CMc3}
    \bar{A}_{ca}^{e_k} = \bar{A}_{ca}[:, e_k], \quad 1 \leq k \leq n.
\end{equation}

Here, $m_k \in \mathbb{R}^{w \times h}$ represents the anchor points mask for concept $e_k$, where $w$ and $h$ represent width and height of latent image respectively. We then use a similar approach to obtain the averaged self-attention map:

\begin{equation}
\label{eq:CMs1}
    \bar{A}_{sa} = \text{MinMaxNorm}\left(\frac{1}{T} \cdot \frac{1}{L} \sum_{t} \sum_{l} A^{l}_{sa(t)}\right).
\end{equation}

Additionally, we apply cross-normalization by subtracting attention maps of other concepts before computing. This ensures that each concept's attention is strongest within its own region and weaker in others:

\begin{equation}
\label{eq:CMs4}
    \bar{A}^{e_k\prime} = \text{MinMaxNorm}\left(\max\left(\bar{A}_{sa}^{e_k} - \frac{1}{n-1} \sum_{e_i \neq e_k} \bar{A}_{sa}^{e_i}, 0\right)\right),
\end{equation}
where 
\begin{equation}
\label{eq:CMs3}
    \bar{A}_{sa}^{e_k} = \bar{A}_{sa}[:, m_k], \quad 1 \leq k \leq n.
\end{equation}

Next, we filter the latent image features to identify region that show high attention to the anchor points, using a relatively low threshold value $s_{sa}$:

\begin{equation}
\label{eq:CMs2}
    d_k[i] = 
    \begin{cases}
    1, & \bar{A}_{sa}^{e_k}[i] \geq s_{sa} \\
    0, & \text{otherwise}
    \end{cases}, 1\leq i \leq w \times h .
\end{equation}

Thus, $\mathbb{D}=\{d_{1}, d_{2}, \dots, d_{n}\}$ represents the concept regions as determined by the diffusion model.

\subsubsection{SP-Attn}
\label{sec:SP-Attn}
To protect concept regions from the influence of irrelevant attributes and concepts, we construct an SP-Mask, indicating which token embeddings should not participate in cross-attention for specific concept regions, which can be formulated as follows:

\begin{equation}
\label{eq:SPA1}
    \begin{cases}
        M_{sp}[d_k][\sum_{i \neq k} a_i + \sum_{i \neq k} e_i] = -\infty, \\
        M_{sp}[d_k][\sim(\sum_{i \neq k} a_i + \sum_{i \neq k} e_i)] = 0, \\
        M_{sp}[\sim(\sum_{k} d_k)][:] = 0,
    \end{cases} \quad 1 \leq k \leq n,
\end{equation}

where $M_{sp} \in \mathbb{R}^{w \times h, l}$ represents the SP-Mask, with $-\infty$ specifying positions of tokens that should not attend in the attention computation.

We then combine the SP-Mask with  Eq.\ref{eq:attention map} to produce an adjusted attention map:

\begin{equation}
  \tilde{A}_t^{(l)} = \text{Softmax}\left(\frac{Q_t^{(l)} K_t^{(l)^{T}} + M_{sp}}{\sqrt{d}}\right).
  \label{eq:sp attention map}
\end{equation}

The positions in $M_{sp}$ set to $-\infty$ result in values of $0$ after applying the softmax function, effectively ensuring that these token positions do not participate in the cross-attention computation.

By multiplying with value matrix and projecting it back to latent image space, we get the semantic correct latent features: 
\begin{equation}
  \phi^{(l)\prime}(z_t) = f_{out}^{(l)} \left(\tilde{A}_t^{(l)} V_t^{(l)}\right).
  \label{eq:hiddenstates}
\end{equation}

\subsubsection{Framework}
SPDiffusion aims to allow the diffusion model to autonomously determine its layout and concept position. Previous work \cite{tumanyan2023plug} shows that layout information is primarily established during the early steps of the denoising process.
Thus, we limit our process to the initial $T_s$ steps rather than performing the entire denoising sequence. During this phase, we save the self and cross attention maps to define concept regions. 
\begin{equation}
    \label{eq:pipeline1}
    z_{t-1}^{*}, \{A_{ca(t)}^{*(l)}\}, \{A_{sa(t)}^{*(l)}\} = \epsilon_{\theta}(z_t^{*}, c(p),t) , (0 \leq t \leq T_s),
\end{equation}

Following Sec.\ref{sec:Concept Region Extraction}, we extract the regions of concepts and construct $M_{sp}$ according Sec.\ref{sec:SP-Attn}. Starting from the same noise, we then proceed with the full semantic protection denoising using $M_{sp}$. To preserve the layout, we replace the self-attention map during the initial $T_s$ steps. This can be formalized as follows:

\begin{equation}
\label{eq:pipeline2}
\begin{split}
z_{t-1} =
\begin{cases}
\epsilon_{\theta}(z_t, {c}(p), M_{sp},t) {A_{sa(t)}^{(l)} \leftarrow A_{sa(t)}^{*(l)}}, & (0 \leq t \leq T_s) \\
\epsilon_{\theta}(z_t, {c}(p), M_{sp},t), & (T_s < t \leq T)
\end{cases}
\end{split}
\end{equation}

Since $T_s$ is typically a small number, SPDiffusion adds minimal inference cost while achieving excellent performance.

%% file: sec/4_expriment.tex
\section{Experiment}

\begin{figure*}[t]
    \centering
    \includegraphics[width=0.9\textwidth]{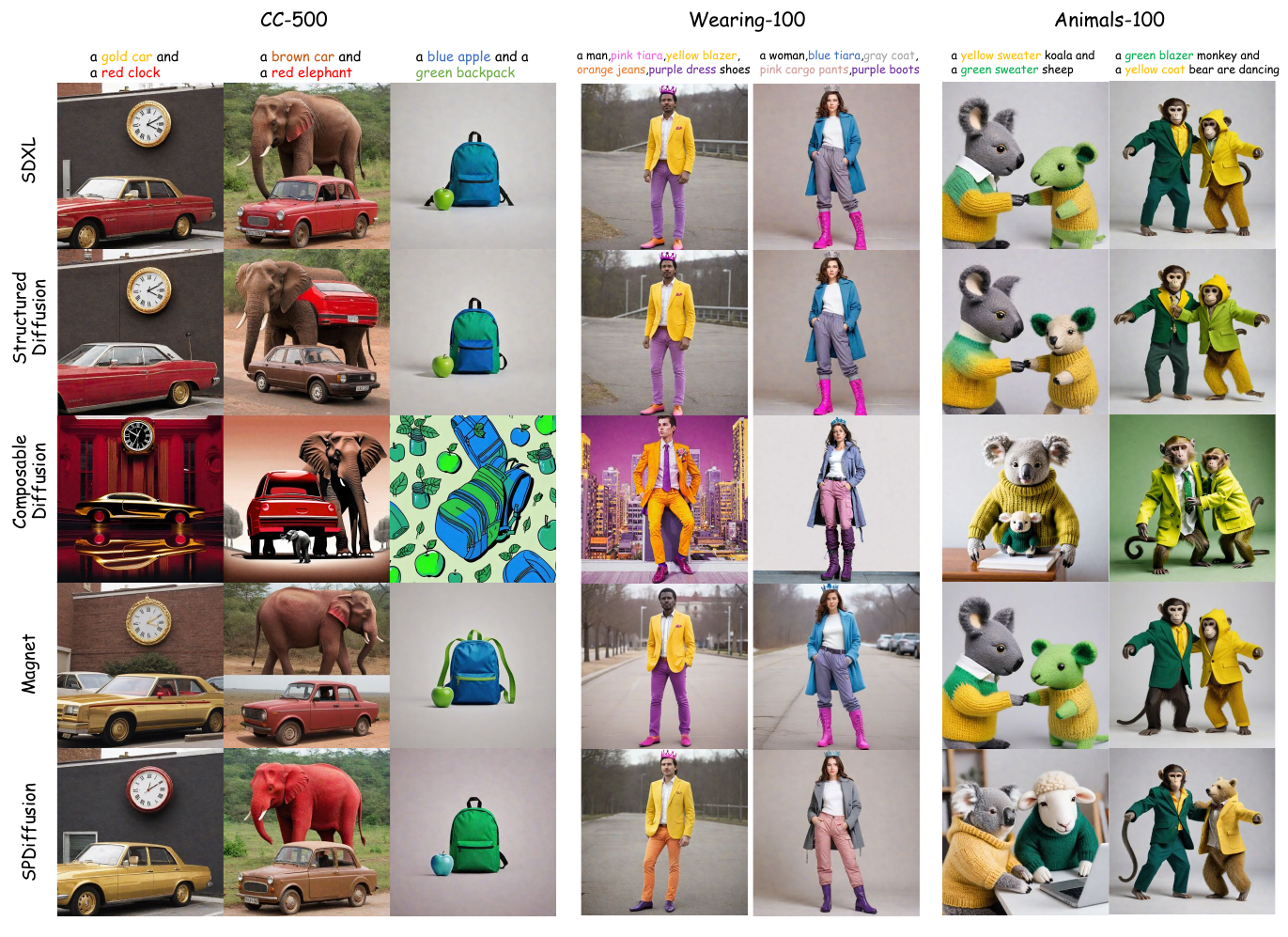} 
    \caption{
    \textbf{Qualitative comparison.} 
    Our method generates address semantic entanglement problems on all three datasets.}
    \label{fig:qualitative_sdxl}
\end{figure*}

\begin{table*}[t]
\small
\centering
\begin{tabular}{lccccccc}
\toprule
\multirow{2}{*}{\textbf{Method}} & \multicolumn{2}{c}{\textbf{CC-500}} & \multicolumn{2}{c}{\textbf{Wearing-100}} & \multicolumn{2}{c}{\textbf{Animals-100}} \\

\cmidrule(lr){2-3} \cmidrule(lr){4-5} \cmidrule(lr){6-7}
& \textbf{BLIP-VQA} $\uparrow$  & \textbf{InternVL-VQA} $\uparrow$ & \textbf{BLIP-VQA} $\uparrow$  & \textbf{InternVL-VQA} $\uparrow$ & \textbf{BLIP-VQA} $\uparrow$  & \textbf{InternVL-VQA} $\uparrow$ \\
\midrule
Stable Diffusion XL\cite{podell2023sdxl}& 0.657& 74.01 & 0.481 & 80.30 & 0.56 & 66.33 \\
Structured Diffusion \cite{feng2022training}& 0.641 & 73.70 & 0.456 & 79.83 & 0.523 & 62.51\\
Composable Diffusion \cite{liu2022compositional}&0.681 & 74.77 & 0.574 &83.76 &0.576 & 61.47\\
Magnet \cite{zhuang2024magnet}& 0.711& 76.73& 0.543 &82.15 & 0.501 & 63.72 \\
\textbf{Ours} &\textbf{0.765} & \textbf{81.57} & \textbf{0.675} & \textbf{87.46} &\textbf{0.631} & \textbf{76.76} & \\
\bottomrule
\end{tabular}
\vspace{-3mm}
\caption{
\textbf{Quantitative Evaluation.} 
Our method outperforms all baselines on all datasets.}
\label{tab:quantitative_sdxl}
\end{table*}

\subsection{Experimental Settings}
\subsubsection{Basic Setups}
Our experiments are primarily conducted on Stable Diffusion XL (SDXL) \cite{podell2023sdxl}. We employ a maximum of 1000 sampling steps, using the DDIM scheduler \cite{song2020denoising} for 20 iterations.
We use classifier-free guidance \cite{ho2022classifier} with a guidance scale of 7.5. 
We use an image size of 768x768. The cross-attention threshold is 0.9 and the self-attention threshold is 0.2. The attention map obtaining and layout maintaining step is 2. SP-Attn is applied in all transformer blocks of Stable Diffusion XL.

\subsubsection{Benchmark} 
\label{sec:benchmarks}
We implement three prompt datasets to evaluate concept disentanglement and attribute binding. For each prompt, we generate 4 images with different seeds during evaluation. The datasets are:

(1). CC-500 \cite{feng2022training}: This dataset contains prompts that combine two concepts, each with one color attribute. The prompt format is: a [color] [subject/object] and a [color] [subject/object]. We randomly sample 100 prompts to maintain consistency with the other two datasets.

(2). Wearing-100: This dataset contains 100 prompts generated with ChatGPT \cite{chatgpt}. Each prompt describes a person wearing four pieces of clothing, each with a distinct color. The format is: a man/woman, [color1] [clothing1], [color2] [clothing2], [color3] [clothing3], [color4] [clothing4].

(3). Animals-100: This dataset contains 100 prompts, also generated with ChatGPT \cite{chatgpt}. Each prompt involves two animals, each with clothing. The format is: a [color] [clothing] [animal] and [color] [clothing] [animal].

\subsubsection{Baseline}
We adopt following training-free method as our baselines:
1). Stable Diffusion XL \cite{podell2023sdxl} 
2). Structured Diffusion \cite{feng2022training} 
3). Composable Diffusion \cite{liu2022compositional} 
4). Magnet \cite{zhuang2024magnet}

\subsubsection{Metric}
We use BLIP-VQA \cite{huang2023t2i} to evaluate the consistency between prompts and generated images. In BLIP-VQA, questions are posed to the BLIP \cite{li2022blip} model regarding each concept and its attributes, if present. The result is a probability indicating the likelihood that the specified concept and attribute exist in the image. The question format is: "[color] [concept]?".

To more precisely measure the consistency between text and images, we also employ the visual large language model InternVL \cite{chen2024internvl} to score the generated images, which we refer to as the InternVL-VQA score. For more details on the InternVL scoring process, please refer to the Supplementary Material.

\subsection{Qualitative Evaluation}
We provide visual comparison images alongside baseline methods, as shown in Fig.~\ref{fig:qualitative_sdxl}. Our method demonstrates strong attribute binding on both CC-500 and Wearing-100, as well as effective concept disentanglement on Animals-100. While baseline methods occasionally manage to bind colors to the correct objects, they generally struggle with concept entanglement issues in Animals-100. Since Structured Diffusion \cite{feng2022training} and Composable Diffusion \cite{liu2022compositional} split text prompt and generate seperately, they often generate disharmonious images. For instance, an elephant’s body is embedded in a red car, as illustrated in the second row and second column.

\subsection{Quantitative Evaluation}
\label{sec:Quantitative Evaluation}
To evaluate the ability to address semantic entanglement more precisely, we conduct quantitative evaluations across all three datasets. Our method outperforms all baseline methods on each dataset in both BLIP-VQA and InternVL-VQA scores, demonstrating superior attribute binding and concept disentanglement. Baseline methods generally score lower than SDXL on Animals-100, indicating limited capability in composing multiple characters within an image. Structured Diffusion, in particular, performs worse across all datasets, likely because replacing concept embeddings increases the gap between individual concept embeddings and the end-of-sentence embedding, which encapsulates the full sentence semantics.

\begin{figure}[t]
    \centering
    \includegraphics[width=0.49\textwidth]{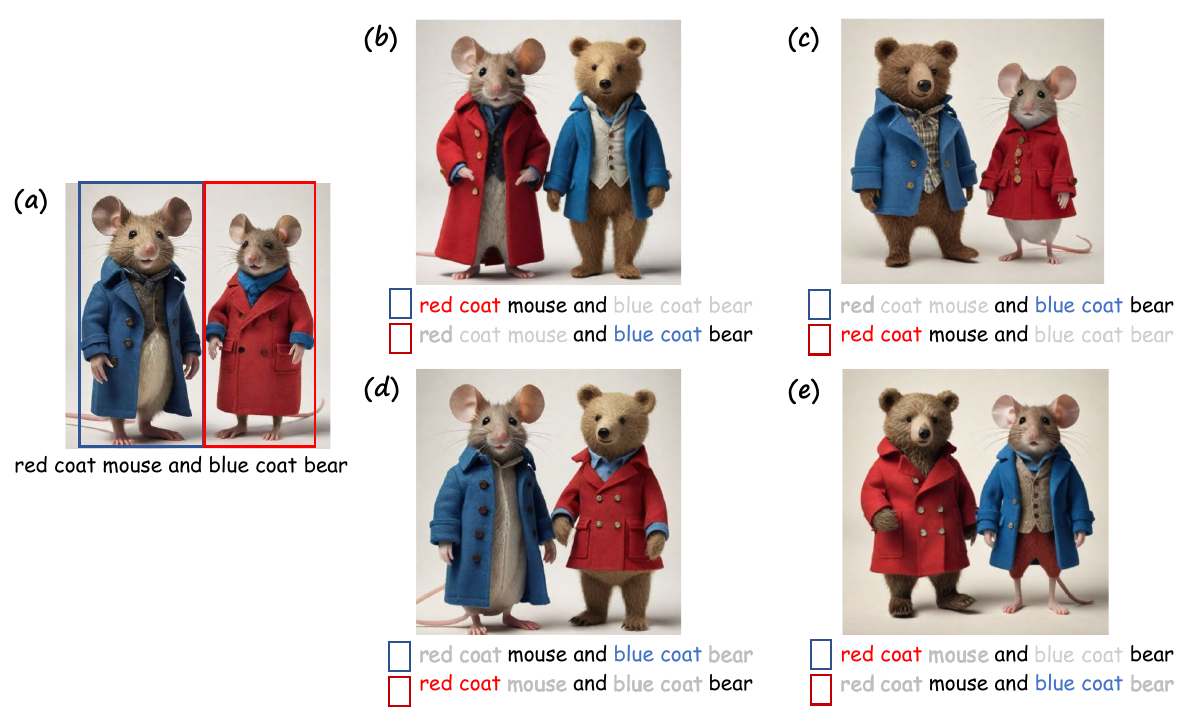} 
    \vspace{-7mm}
    \caption{
    \textbf{Qualitative Ablation for Different Tokens.}
    By masking different tokens for different regions, we can manipulate the attention of certain regions for specific tokens.}
    \label{fig:ablation_for_tokens} 
\end{figure}

\begin{figure}[t]
    \centering
    \includegraphics[width=0.43\textwidth]{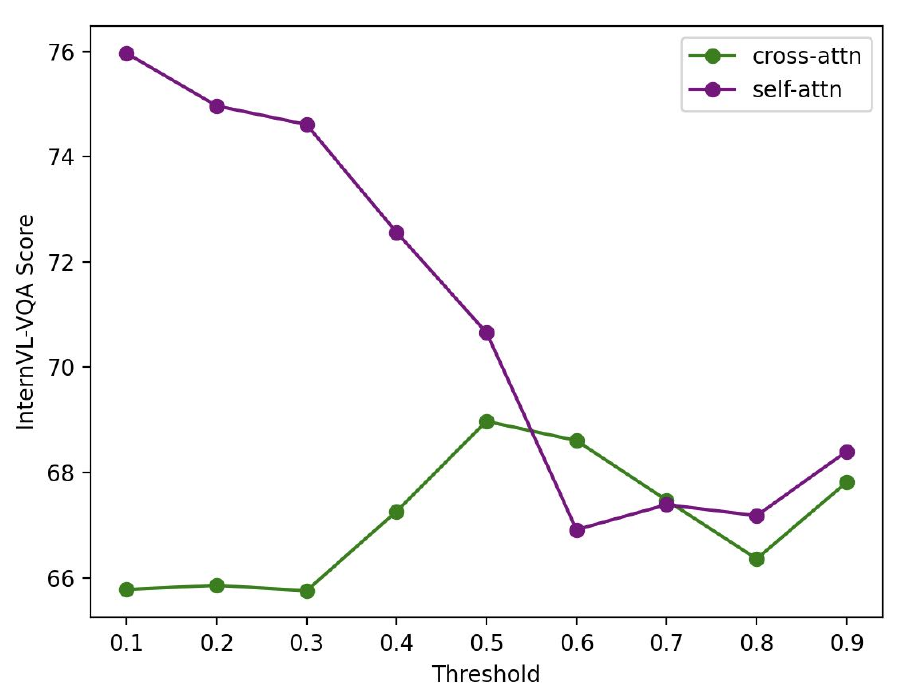} 
    \caption{
    \textbf{Ablation for different threshold and region extraction method}. 
    SP-Extraction method outperforms the method directly getting concept region from cross-attention map. 
    }
    \label{fig:ablation_for_threshold} 
\end{figure}


\subsection{Ablation Study}
\label{sec:ablation_study}
Since the attribute and concept tokens are closely packed together in the evaluation datasets, we further demonstrate the effectiveness of our method by conducting an ablation study on different mask settings for these tokens. 
As shown in Fig.~\ref{fig:ablation_for_tokens} (a), the absence of semantic protection results in both sides depicting mouse-like images, indicating that the mouse token receives high attention on both sides.
In Fig.~\ref{fig:ablation_for_tokens} (b), we mask the ``blue coat bear'' tokens for blue box region and the ``red coat mouse'' tokens for red box region. This adjustment successfully corrects the appearance of the bear and the mouse, as well as their respective clothing colors.
Similarly, by swapping the token groups masked in blue and red box region, we can switch the positions of the characters, as shown in Fig.~\ref{fig:ablation_for_tokens} (c). 
Furthermore, by masking different clothing and color tokens, we can determine the clothing colors of the characters, as illustrated in Fig.~\ref{fig:ablation_for_tokens} (d)(e). 
This experiment shows that incorrect attention to certain tokens in the latent features leads to semantic entanglement. By protecting specific regions in the latent features from irrelevant tokens, we can restore the intended semantics and correct these errors.

We also conduct ablation studies to evaluate the process and locations where concept regions are obtained. In this study, we apply different thresholds to filter concept regions directly from the cross-attention map and compare the results with our SP-Extraction method. The SP-Extraction method first identifies anchor points in the cross-attention map for each concept, and then obtains the final concept mask using the self-attention map. We test the methods on the Animals-100 dataset to assess performance in scenarios with concept entanglement. We set the anchor point threshold at 0.9. As shown in Fig.~\ref{fig:ablation_for_threshold}, directly extracting concept regions from the cross-attention map achieves its peak performance at a threshold of 0.5, with an InternVL-VQA score of no more than 70. In contrast, the SP-Extraction method achieves its best performance with a cross-attention threshold of 0.9 and a self-attention threshold of 0.1, resulting in an InternVL-VQA score of 76.6. Additionally, within the threshold range of 0.1 to 0.5, the SP-Extraction method consistently outperforms the direct cross-attention approach, demonstrating its robustness and insensitivity to variations in threshold values.

%% file: sec/5_conclusion.tex
\section{Conclusion}
In this work, we propose the Semantic Protection Diffusion (SPDiffusion) to handle the semantic entanglement problem in multi-concept text-to-image generation. SPDiffusion utilizes SP-Extratction to extract concept region from a incorrect image generation. It utilizes a SP-Mask to indicate the relevance of the regions and the tokens, and design a SP-Attn to shield the influence of irrelevant tokens on specific regions in the generation process. We conduct extensive experiments and demonstrate the effectiveness of our approach, showing advantages over other methods in both attribute binding and concept disentanglement. 
We believe our method and insight can support further development for solving semantic entanglement problems in multi-concept generation.

%% file: sec/X_suppl.tex
\clearpage
\setcounter{page}{1}
\maketitlesupplementary

\section{InternVL-VQA}
In this paragraph, we will introduce the InternVL-VQA score in detail. We use a Visual Language Model, InternVL 2, to score the generated images to evaluate the alignment between the images and the text prompts. We conduct two rounds of questions and answers. In the first round, we ask InternVL to describe the content of the image. In the second round, we ask InternVL to score the alignment between the image and the text prompt on a scale from 0 to 100. The questions are shown below:

\begin{enumerate}
    \item You are my assistant to identify the animals or objects in the image and their attributes. Briefly describe the image within 50 words.
    \item According to the image and your previous answer, evaluate how well the image aligns with the text prompt: \{prompt\}.
    100: the image perfectly matches the content of the text prompt, with no discrepancies.\
                    80: the image portrayed most of the actions, events and relationships but with minor discrepancies.\
                        60: the image depicted some elements in the text prompt, but ignored some key parts or details.\
                            40: the image did not depict any actions or events that match the text.\
                                20: the image failed to convey the full scope in the text prompt.\
                                    Provide your analysis and explanation in JSON format with the following keys: explanation (within 20 words),score (e.g., 85)." 
\end{enumerate}

\section{Application}
Our method can be applied to any scenario where cross-attention is involved and the depiction of multi-character is suboptimal, such as in ControlNet \cite{controlnet}, StoryDiffusion \cite{zhou2024storydiffusion}, and PhotoMaker \cite{li2024photomaker}.

\begin{figure*}[t]
    \centering
    \includegraphics[width=0.8\textwidth]{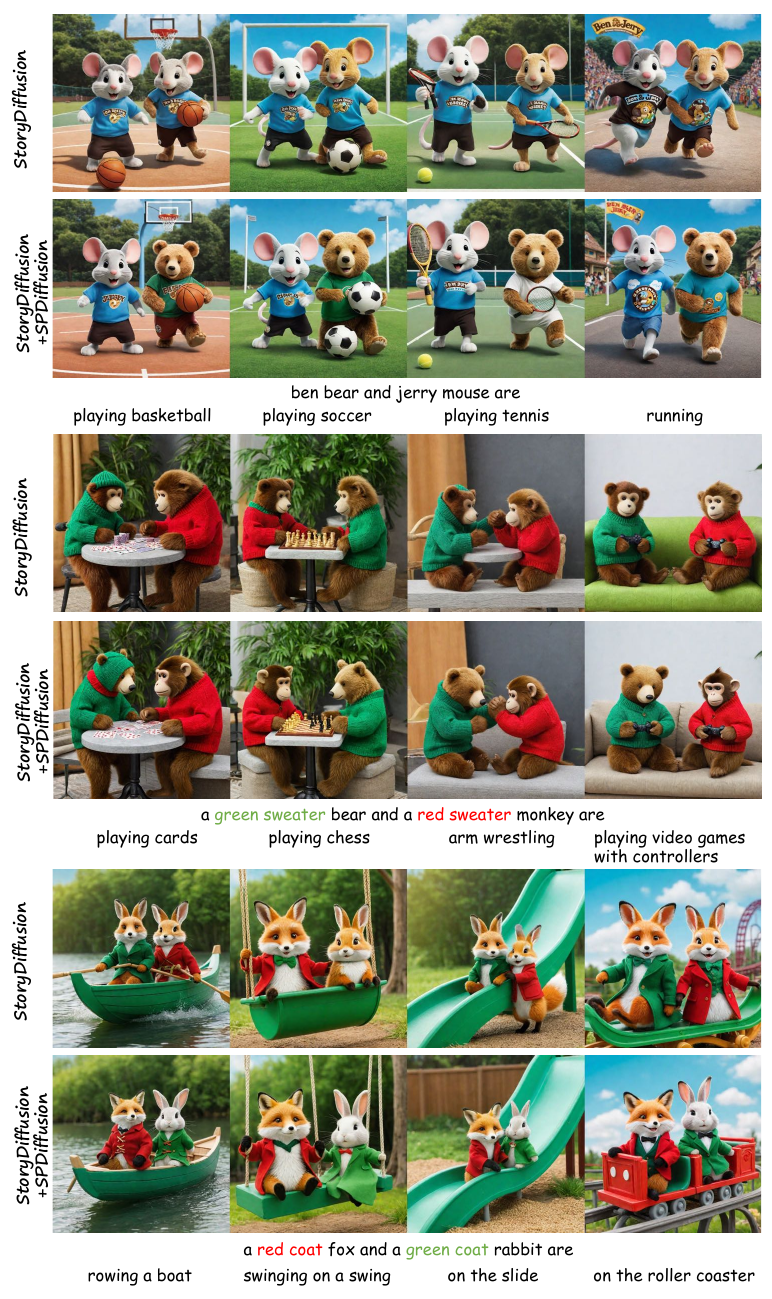} 
    \caption{
        Our method can be integrated with StoryDiffusion to enhance the storytelling capabilities of multi-character generation.
    }
    \label{fig:additional_storydiffusion} 
\end{figure*}

StoryDiffusion is designed to ensure the consistency of character images throughout a generated sequence, primarily using self-attention. Our method, which focuses on cross-attention, can be seamlessly integrated with StoryDiffusion to achieve consistency of multiple characters across consecutive frames in a story, as demonstrated in Fig.~\ref{fig:additional_storydiffusion}.

\begin{figure*}[htbp]
    \centering
    \includegraphics[width=1\textwidth]{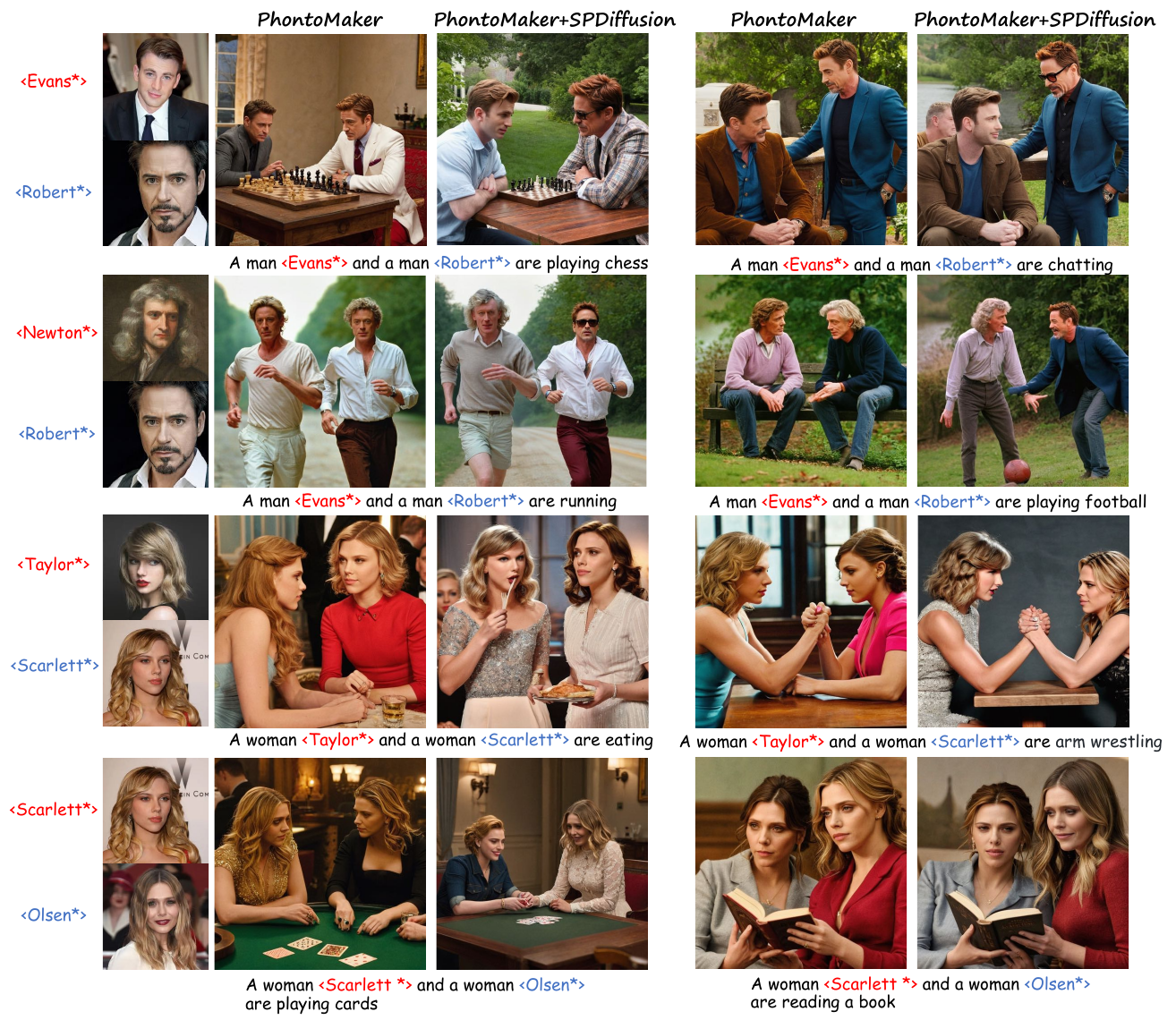} 
    \caption{
        Our method can be integrated with PhotoMaker to enhance the performance of multi-character generation.
    }
    \label{fig:additional_photomaker} 
\end{figure*}

PhotoMaker generates character images based on provided reference images, maintaining character identity by embedding character features into class tokens. However, when two or more different characters appear, their appearances may fuse. Our method effectively separates the appearances of the characters, preserving their distinct identities, as shown in Fig.~\ref{fig:additional_photomaker}. This demonstrates that our method can be applied to any multi-character generation scenario based on character tokens, showcasing strong versatility.

\section{Additional Qualitative Results}
\label{sec:Additional Qualitative Results}
\begin{figure*}[htbp]
    \centering
    \includegraphics[width=0.8\textwidth]{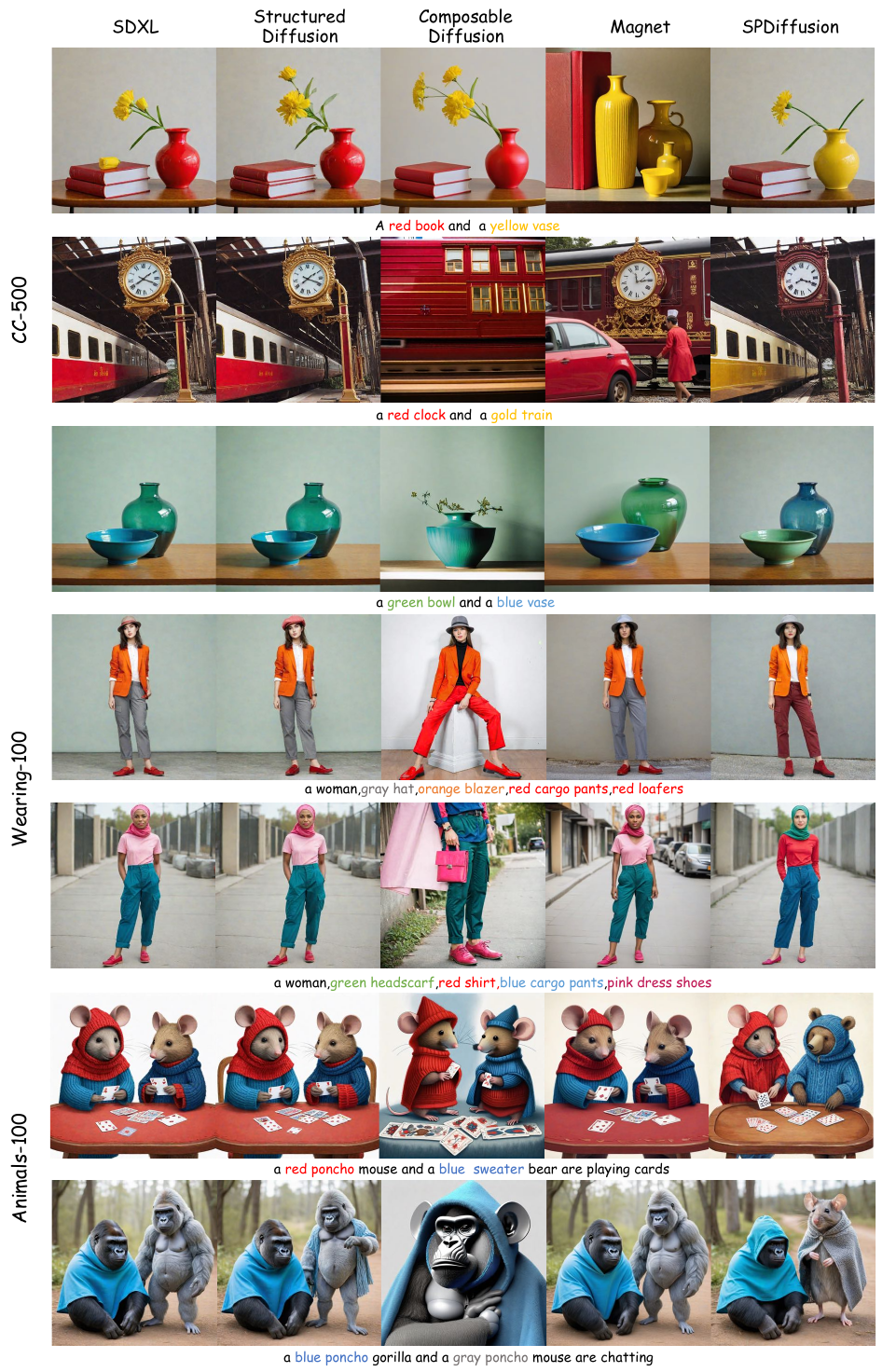} 
    \vspace{-10pt}
    \caption{
        Additional Qualitative Results.
        }
    \label{fig:additional_results1} 
\end{figure*}

\begin{figure*}[htbp]
    \centering
    \includegraphics[width=0.8\textwidth]{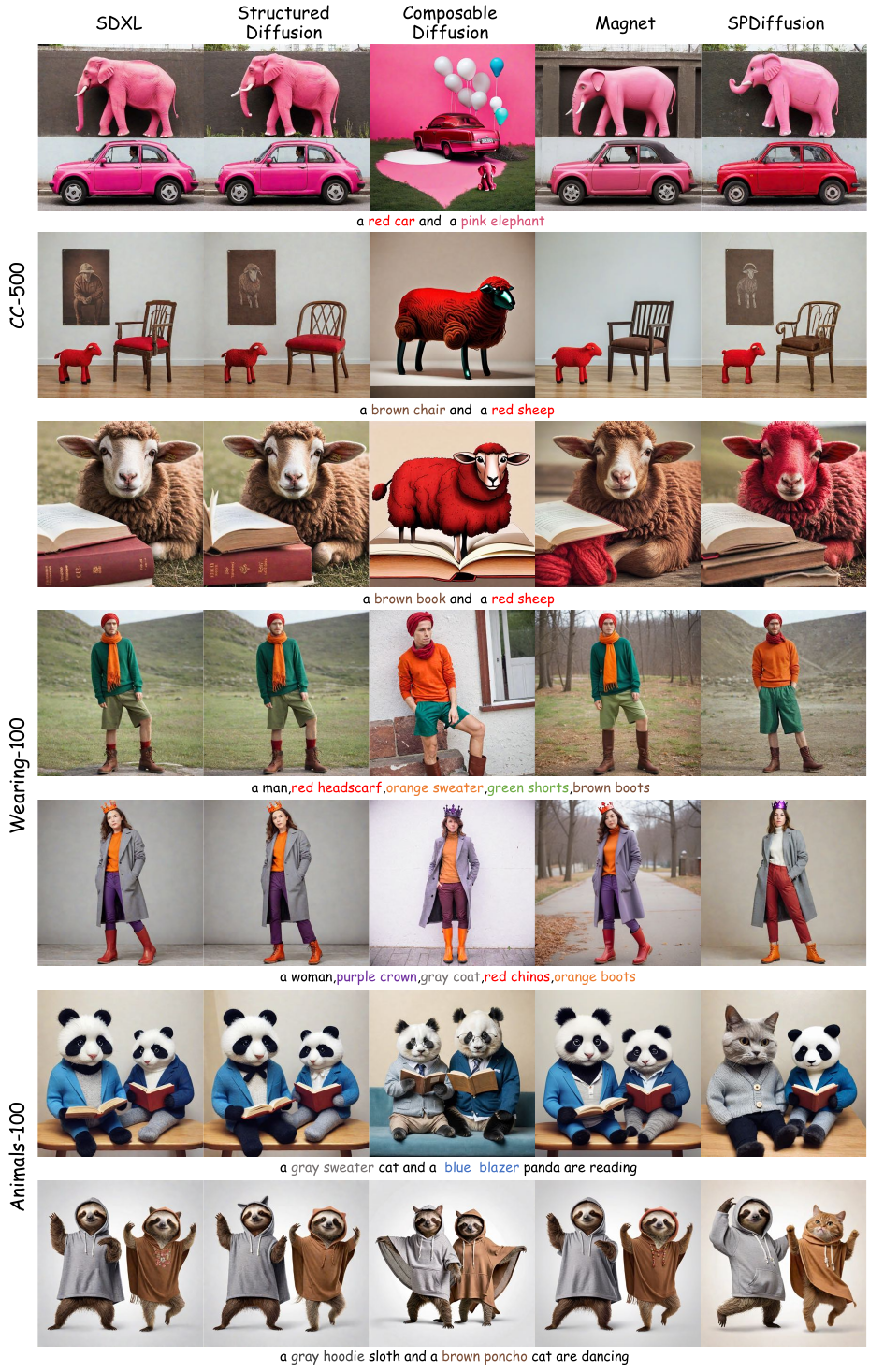} 
    \vspace{-10pt}
    \caption{
        Additional Qualitative Results.
        }
    \label{fig:additional_results2} 
\end{figure*}

We provide additional qualitative comparisons between the baseline methods and our method across all three datasets, as shown in Fig.\ref{fig:additional_results1} and Fig.\ref{fig:additional_results2}.

